\documentclass{article}
\usepackage{ijcai23}

\usepackage{times}
\usepackage{soul}
\usepackage{url}
\usepackage[hidelinks]{hyperref}
\usepackage[utf8]{inputenc}
\usepackage[small]{caption}
\usepackage{graphicx}
\usepackage{amsmath}
\usepackage{amssymb}
\usepackage{amsthm}
\usepackage{booktabs}
\usepackage{algorithm}
\usepackage{algorithmic}
\usepackage[switch]{lineno}
\usepackage[11pt]{moresize}
\usepackage{xcolor}

\DeclareMathOperator*{\argmin}{arg\,min}

\title{Bayesian Federated Learning: A Survey}

\author{
Longbing Cao$^{1,2}$\and
Hui Chen$^1$\and
Xuhui Fan$^{3}$\and
Joao Gama$^4$\and
Yew-Soon Ong$^5$\And
Vipin Kumar$^6$
\affiliations
$^1$University of Technology Sydney, Australia\\
$^2$Macquarie University, Australia\\
$^3$University of New Castle, Australia\\
$^4$University of Porto, Portugal\\
$^5$Nanyang Technological University, Singapore\\
$^6$University of Minnesota, USA
\emails
}

\begin{document}

\maketitle

\begin{abstract}
Federated learning (FL) demonstrates its advantages in integrating distributed infrastructure, communication, computing and learning in a privacy-preserving manner. However, the robustness and capabilities of existing FL methods are challenged by limited and dynamic data and conditions, complexities including heterogeneities and uncertainties, and analytical explainability. \textit{Bayesian federated learning} (BFL) has emerged as a promising approach to address these issues. This survey presents a critical overview of BFL, including its basic concepts, its relations to Bayesian learning in the context of FL, and a taxonomy of BFL from both Bayesian and federated perspectives. We categorize and discuss client- and server-side and FL-based BFL methods and their pros and cons. The limitations of the existing BFL methods and the future directions of BFL research further address the intricate requirements of real-life FL applications. 
\end{abstract}


\section{Introduction}

Decentralized AI (DeAI) and machine learning (DeML) address new demands for engaging decentralized edge devices, nodes and servers to undertake secure edge-level and distributed AI and learning tasks \cite{cao2022decentralized,mcmahan2017communication}. A critical DeAI and ML technique is federated learning (FL) \cite{kairouz2021advances,Yang21}, which ensures decentralized learning on local data without sharing but with properties including privacy, security, heterogeneity, and personalization. FL has seen significant developments to leverage distributed and centralized ML techniques, catering to centralized, decentralized, heterogeneous, or personalized settings and requirements with privacy preservation. These include algorithms for preserving privacy \cite{liu2022on,elgabli2022fednew}, reducing communication consumption \cite{zhang2021fedpd}, handling system heterogeneity \cite{zong2021communication}, and addressing personalized requirements \cite{t2020personalized,fallah2020personalized}.

However, FL faces various fundamental challenges in enabling DeAI and DeML for real-world applications. First, client data on edge nodes and devices may be very limited and it may be costly to obtain labelled samples. Second, client conditions and behaviors are often dynamic, presenting strong uncertainties. Last but not least, decentralized applications are non-IID, involving heterogeneities of devices, behaviors, goals, and data and their interactions \cite{cao2022beyond}. These are challenges facing existing FL algorithms and inspire a promising direction - Bayesian Federated Learning (BFL) \cite{zhang2022personalized}, which integrates the advantages of Bayesian learning (BL) into FL. 
In existing BFL methods, BL approaches incorporate prior knowledge about data to leverage a limited number of samples and learn distributions over FL parameters and statistical heterogeneity to quantify uncertainties and dynamics. Further, the strengths of FL in handling privacy, communication and heterogeneity are fused with BL. Consequently, BFL enables more robust, well-calibrated and dynamic predictions for safety- and privacy-critical applications. BFL has demonstrated various applications, such as for FinTech, driverless cars, Industry 4.0, medical diagnosis, and differential privacy \cite{achituve2021personalized,snell2021bayesian,kendall2017uncertainties,blundell2015weight,TriastcynF19,Yang21}. 

BFL shows a promising potential to substantially expand the existing FL and complex real-world requirements for DeAI, DeML and FL. Here, we systematically review the relevant work on BFL, conduct a critical analysis of their pros and cons, and present the challenges and opportunities for comprehensive BFL research. First, we briefly discuss the transfer from FL and BL to their integrative BFL. Then, a BFL taxonomy offers a structure of BFL in terms of both BL methods and FL research issues. We discuss and compare the advantages and disadvantages of different methods for client- and server-side BFL and various categories of BFL from the FL perspective. Lastly, we discuss the gaps and future directions of BFL in addressing broad-reaching, more realistic and actionable \cite{pakdd_CaoZ06,widm_Cao13} FL, BL, DeAI, and DeML settings and tasks and real-world FL scenarios and requirements.

\section{From Federated and Bayesian Learning to Bayesian Federated Learning}

We summarize FL concepts, topics and challenges. Then, BL methods and advantages are discussed. BFL integrates BL into FL addressing FL challenges using BL advantages.

\subsection{Federated Learning and Challenges}
\label{subsec:fl}

\textbf{FL concepts}. Federated learning initially aims to address privacy leakage in  distributed learning systems \cite{Yang21,kairouz2021advances}. In general, an FL system has a server and multiple clients and utilizes an iterative learning process through server-client communications. In each communication round, the server trains a global model and sends its model parameters to participating clients. Each client trains its local model on local data. Clients share model parameters (rather than local data) with the server, which pools and aggregates these local updates to update its global model for the next communication round. This process iterates until convergence or it reaches some stopping conditions. 

Typically, FL achieves an objective function as follows:
\begin{equation}
\label{eq:generalFL}
    \min _{\boldsymbol{w}} F(\boldsymbol{w})=\sum_{m=1}^M \frac{\left|D_m\right|}{|D|} F_m(\boldsymbol{w}), 
\end{equation}
\begin{equation}
\label{equation 2}
    \quad \text { where } \quad F_m(\boldsymbol{w})=\frac{1}{\left|D_m\right|} \sum_i l\left(\boldsymbol{x}_i, y_i ; \boldsymbol{w}\right).
\end{equation}
Here, $M$ denotes the number of activated clients, $D$ and $D_m$ denote the aggregated dataset from all participating clients and the dataset of the $m$-th client, respectively, $\boldsymbol{w}$ denotes the model parameters of Bayesian neural networks and  $F_m(\boldsymbol{w})$ denotes the empirical risk of the $m$-th client with the loss function $l$ over each instance $(\boldsymbol{x}_i,y_i)$ \cite{chen2022on}. 

\textbf{FL research topics}. FL has seen diversified rapid developments, which include the following categories. (1) Privacy-preserving FL \cite{liu2022on,elgabli2022fednew} for differential privacy; (2) Communication-efficient FL \cite{lin2022personalized,li2022soteriafl} addressing communication bottlenecks, \cite{dinh2020federated} on resource allocation, \cite{zang2022traffic} on sparsified and compressed communication, and \cite{liu2020client} for propagating channels; (3) Heterogeneous and personalized FL \cite{ghari2022personalized,t2020personalized} to address heterogeneities of local clients; and (4) FL optimization with different settings \cite{liu2021bayesian,malinovskiy20a} for federated SGD, \cite{mcmahan2017communication} for federated averaging, \cite{chen2021dynamic} for dynamic aggregation, and \cite{durmus2021federated} for dynamic regularization. In addition, there are many other (1) FL tasks, such as hierarchical FL, non-IID FL, unsupervised to semi-supervised FL, multitask FL, robust FL, fair and unbiased FL; and (2) FL application settings, such as blockchained FL, multimodal FL, secure FL, energy-aware and green FL.    

\textbf{FL challenges}. However, these previous work on FL still faces various challenges as mentioned in the introduction. In particular, FL suffers from uncertain, dynamic, limited, and unsupervised problems and also data. These techniques also cannot provide robust and analytically explainable results.

\subsection{Bayesian Learning and Advantages}

\textbf{BL concepts}. Bayesian learning (BL) builds on Bayes' theorem \cite{BenavoliC21,SunCL20}.
It aims at learning the posterior distribution $p(\boldsymbol{w} \mid D)$ for parameter $\boldsymbol{w}$ based on its prior distribution $p(\boldsymbol{w})$ and the corresponding likelihood $p(D\mid \boldsymbol{w})$ on observed evidence $D$. 
BL then learns the posterior as follows:
\begin{equation}
    p(\boldsymbol{w} \mid D)=\frac{p(D\mid \boldsymbol{w}) p(\boldsymbol{w})}{p(D)},
\end{equation}
It thus converts a prior probability $p(\boldsymbol{w})$ into a posterior probability $p(\boldsymbol{w} \mid D)$ with the likelihood $p(D\mid \boldsymbol{w})$ for parameter $\boldsymbol{w}$ on observed evidence $D$.

\textbf{BL methods}. Bayes' theorem has been widely applied in various settings and has generated numerous BL methods \cite{jospin2022hands,gershman2012tutorial} for tasks such as Bayesian classification, regression, clustering, representation, and optimization. In addition, these can be categorized in terms of modeling mechanisms and settings, learning tasks, and application scenarios, for example (1) Bayesian classification, such as naive Bayes and Bayesian belief networks for classification or optimization; (2) Bayesian approximation, inference and optimization, such as Laplace's approximation, expectation propagation, and variational approximation and Markov chain Monte Carlo (MCMC) methods such as Gibbs sampling; (3) hierarchical Bayesian models with hierarchical variable/parameter dependence, such as Beta-Poisson models; (4) dynamic BL such as Bayesian nonparametric models; (5) BL for other settings and tasks, such as Bayesian continual learning, Bayesian model averaging, and Bayesian posterior decomposition; and (6) hybrid BL methods, such as Bayesian neural networks, and Bayesian model ensemble.


\textbf{BL advantages}. BL shows various advantages. (1) \textit{Quantifying uncertainty} over probability distributions rather than specific values, which captures the model or epistemic uncertainty of model parameters \cite{kendall2017uncertainties}. This can justify the reliability of results, such as for safety-critical applications. (2) \textit{Enhancing robustness} by evidence-based likelihood estimates, which learns the parameter distributions and regularizes the model for more robust parameters. (3) \textit{Improving performance on limited data} using prior distribution for each model parameter, which captures prior knowledge about the data and problem. This can greatly improve modeling performance particularly with limited data \cite{sun2019functional}. 

\subsection{BFL: Bayesian Federated Learning}

The discussed challenges facing FL can be mostly overcome by taking advantage of BL to address various real-life needs. Consequently, Bayesian federated learning (BFL) has been explored to incorporate BL principles and advantages into FL frameworks and tasks for stronger model robustness and learning improved performance on small-scale data \cite{zhang2022personalized}. BFL could lead to more robust, better explainable, and higher performance in handling uncertainties and process-oriented (rather than point-based) challenges. Such advantages could benefit various applications with strong uncertainties, such as estimating financial market dynamics, medical conditions, infectious diseases, financial crisis, and natural disaster.

\begin{figure}[tb!]
    \centering
    \includegraphics[width=0.8\columnwidth]{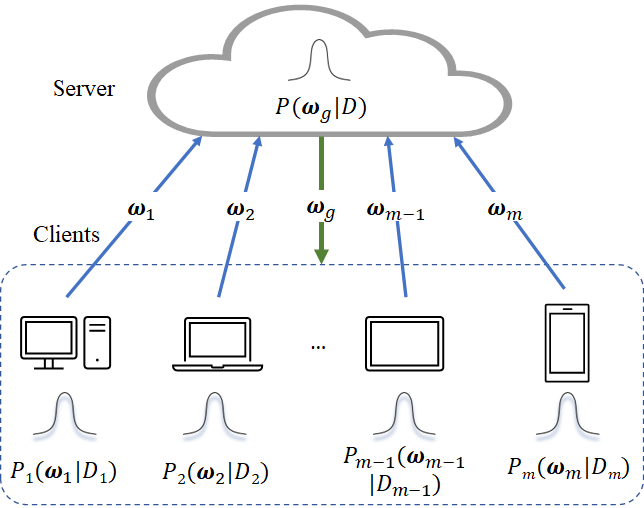}
    \caption{The framework and iterative learning process of Bayesian federated learning (BFL).}
    \label{Fig:BFL-process}
\end{figure}

Although no unified definitions are available for BFL, Figure \ref{Fig:BFL-process} illustrates (1) the general framework of BFL, which integrates mechanisms of FL and BL; and (2) a general iterative learning process of BFL. Differing from pure FL, a BFL system learns global posterior $p(\boldsymbol{w} \mid D)$ for the server and local posterior $p_m(\boldsymbol{w}_m \mid D_m)$ for each participating client $m$ on its local data $D_m$. A general or localized prior $p(\boldsymbol{w})$ applies to all clients. 
BFL incorporates Bayesian principles into FL for the server: 
\begin{equation}
    p(\boldsymbol{w}_{g} \mid D)=\frac{p(D\mid \boldsymbol{w}_{g}) p(\boldsymbol{w}_{g})}{p(D)},
\end{equation}
and clients:
\begin{equation}
    p(\boldsymbol{w}_m \mid D_m)=\frac{p(D_m\mid \boldsymbol{w}_m) p(\boldsymbol{w}_m)}{p(D_m)}.
\end{equation}
$\boldsymbol{w}_{g}$ and $\boldsymbol{w}_m$ represent the parameters of the server and the $m$-th client respectively, $p(D)$ denotes a normalization constant. In non-personalized BFL, $\boldsymbol{w}_{g}$ and $\boldsymbol{w}_m$ are equivalent. 

Accordingly, a BFL model converts the FL objective function Eq. (\ref{eq:generalFL}) to BFL for a global loss, aligned local losses, or a mixed loss with settings (e.g., privacy or security preservation, or communication efficiency) and regularization:
\begin{equation}\label{eq:bfl-obj}
    \argmin_{\boldsymbol{w}_m, \boldsymbol{w}_{g}} \sum_{m=1}^{M} l(\boldsymbol{w}_m; D_m) + \alpha \| \boldsymbol{w}_m, \boldsymbol{w}_g \|.
\end{equation}
Then, the posteriors $p(\boldsymbol{w}|D)$ can be approximated by Bayesian optimization, for example, variational inference (VI). Accordingly, Eq. \eqref{eq:bfl-obj} can be converted to an ELBO-like objective $\mathcal{L}_D$, the true posterior $p(\boldsymbol{w}_m, \boldsymbol{w}_g|D)$ is approximated by a variational distribution $q_{\theta}(\boldsymbol{w}_m, \boldsymbol{w}_g)$ estimated on the evidence $D$ with parameters $\theta$: 
\begin{equation}
    \argmin_{\boldsymbol{w}_m, q_{\theta}(\boldsymbol{w}_m)} KL(q_{\theta}(\boldsymbol{w}_m) \| p(\boldsymbol{w}_m)) - \mathbb{E}_{q_{\theta}(\boldsymbol{w}_m)}[\log p(D_m|\boldsymbol{w}_m)]. 
\end{equation}
KL refers to the Kullback–Leibler divergence. Such an objective function determines the optimization of a BFL system in fulfilling client or global objectives, or their mixture; and distinguishes generic FL from personalized FL.

\section{BFL Taxonomy and Analyses}
\label{sec:BFLtax-anal}

Here, we present a taxonomy of BFL and conduct critical analyses of representative BFL methods from both Bayesian and federated perspectives.

\subsection{BFL Taxonomy}

\begin{figure*}[htbp]
    \centering
    \includegraphics[width=2.0\columnwidth]{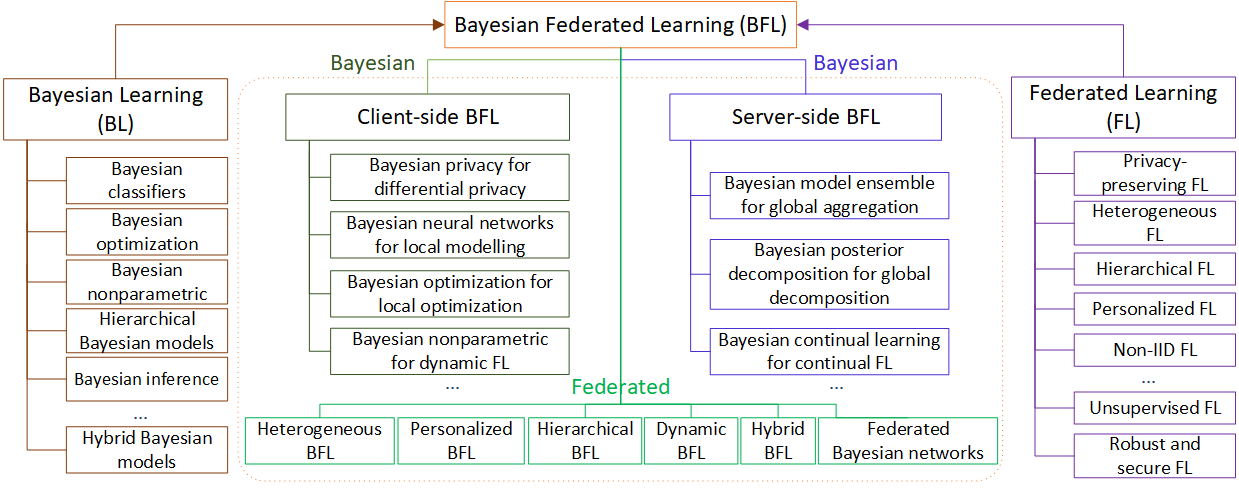}
    \caption{Taxonomy of Bayesian federated learning (BFL) and its connections to Bayesian learning and federated learning. The right and left panels show the mechanisms of FL and BL to support BFL, respectively. The middle panel consists of BFL methods in terms of Bayesian (client- and server-side) and federated perspectives.}
    \label{fig:bfl-taxonomy}
\end{figure*}

BFL tasks and methods can be categorized per aspects such as FL settings or BL methods. First, taking a Bayesian perspective, we categorize BFL in terms of FL architectures: client-side BFL and server-side BFL. Client-side BFL focuses on learning local models on client nodes using Bayesian methods, which involves representative methods: (1) Federated Bayesian  privacy (FBP), (2) Bayesian neural networks (BNNs) for local models, (3) Bayesian optimization (BO) for local optimization, and (4) Bayesian nonparametric (BNP) models for dynamic FL. Server-side BFL aggregates local updates for global models using Bayesian methods, with typical methods including (1) Bayesian model ensemble (BME) for aggregation, (2) Bayesian posterior decomposition (BPD), and (3) Bayesian continual learning (BCL). Then, from the FL perspective, we can categorize BFL methods into heterogeneous, hierarchical, dynamic, personalized, and hybrid BFL, etc. 
Figure \ref{fig:bfl-taxonomy} shows a taxonomy of BFL and its connections to BL and FL, respectively.

\subsection{Client-side BFL}

Various Bayesian methods address client requirements and learn local models. Representative client-side methods include FBP, BNN, BO and BNP for different client-side requirements and objectives.

\subsubsection{Federated Bayesian Privacy}
Earlier FL models focus on privacy-preserving federated updating, communication, and aggregation, with typical progress on differential privacy \cite{elgabli2022fednew}. Various differential privacy methods involve simple statistics and object or output perturbation. By considering the randomness of local data, Bayesian differential privacy (BDP) \cite{TriastcynF19} ensures client privacy, instance privacy and their joint privacy by a privacy loss accounting method. In \cite{GuBP19}, KL-divergence quantifies Bayesian privacy loss during data restoration, forming a federated deep learning for private passport (FDL-PP) method against FL restoration attacks. 

\textbf{Discussion}. Both BDP and FDL-PP relax constraints on existing FL differential privacy by incorporating uncertainty. However, they cannot handle complex FL and BL settings and privacy-preserving requirements.

\subsubsection{Bayesian Neural Networks for FL Local Models}

Bayesian neural networks (BNNs) \cite{blundell2015weight} combine Bayesian inference with neural networks. 
BNNs are incorporated into FL in various ways. pFedBayes \cite{zhang2022personalized} uses a BNN to train local models in each communication round. Its objective is formulated as a two-level optimization problem. For each client, pFedBayes uses VI to approximate the posterior distribution of local model parameters by minimizing the loss function of the local model. On the server, pFedBayes aggregates local models and minimizes the averaged loss functions of clients. pFedBayes \cite{kairouz2021advances} updates local parameters again after updating them using BNN for each client, resulting in a localized global model to overcome the challenge of non-IID data. FedPPD \cite{bhatt2022bayesian} also applies a BNN to train local models. The key difference between FedPPD and pFedBayes lies in the methods of approximating posterior distributions of model parameters. pFedBayes uses VI while FedPPD uses MCMC. FedPPD utilizes a Bayesian dark knowledge method to distill the posterior prediction distribution into a single deep neural network (DNN) for each client and then sends the resulting teacher model (approximate maximum posterior sample) and student model (approximate posterior predictive distribution) to the server for aggregation.

\textbf{Discussion}. BNNs train local models and enable FL to quantify local uncertainty while using DNNs for task learning, thus improving FL robustness.  BNNs also improve the FL learning performance on limited data. However, BNNs also bring challenges to FL. First, the BFL with BNNs involves huge computational and memory costs for training local models, especially when the scale of local model parameters is immense. Second, it may be challenging to choose appropriate prior distributions for local model parameters particularly when we cannot estimate complex relationships between model outputs and parameters.

\subsubsection{Bayesian Optimization for FL Local Optimization}

Bayesian optimization (BO) is a sequential optimization approach, 
often used to tune the hyperparameters of DNNs. Various methods involve BO for FL. In \cite{dai2020federated}, a federated Bayesian optimization (FBO) setting uses BO to optimize local models in each communication round. They propose the algorithm FTS for FBO, where a Gaussian process (GP) is used as the surrogate for modelling the objective and acquisition functions by Thompson sampling. Random Fourier features are used to approximate GP for scalability and information exchange (the parameters of random Fourier features are shared in each communication round). FTS has a strict convergence guarantee even for non-IID data. In \cite{zang2022traffic}, BO supports FL for traffic flow prediction (TFP), and like FTS and other BO algorithms, GP is used as an ideal objective function surrogate. Unlike FTS, since  FBO for TFP utilizes  BO dynamically to adjust the weights of local models for aggregation,  FBO for TFP does not suffer from performance degradation when encountering heterogeneous data.  Moreover, FBO for TFP involves an expected improvement instead of Thompson sampling as the acquisition function of the models without addressing the scalability problem of GP.

\textbf{Discussion}. Applying BO to FL achieves a relatively robust learning performance for non-IID local datasets due to the inherent properties of BO. Compared with traditional optimization algorithms, BO is simpler and more convenient to implement. However, the practicality of FBO is challenging for models with substantial data points. Moreover, the slow convergence rate of FBO is open to address. 

\subsubsection{Bayesian Nonparametric models for Dynamic FL}

Bayesian nonparametric (BNP) models enable dynamic learning \cite{gershman2012tutorial}. 
In BFL, BNP models apply a Gaussian or Beta-Bernoulli process (BBP). In GP-based FL, pFedGP \cite{achituve2021personalized} utilizes a personalized GP classifier to train a local model for each client and shares a kernel function for all clients. FedLoc \cite{yin2020fedloc} also uses GP to train local models for regression tasks. Unlike pFedGP, FedLoc cannot effectively deal with non-IID data for FL \cite{li2020federated}. Different from these methods, FedCor \cite{tang2022fedcor} uses GP to predict the loss change and then selects the clients that need to be activated in each communication round according to the loss change. FedCor is only applicable for the cross-silo FL framework, where the learning performance decays significantly for non-IID data.

In BBP-based FL, PFNM \cite{yurochkin2019bayesian} uses BBP to find the matched subsets of neurons among local models. However, PFNM is only applicable to a simple feedforward neural network structure. Then, a layer-wise matching algorithm FedMA  based on BBP extends PFNM to other neural network structures \cite{wang2020federated} such as convolutional neural networks (CNNs) and long short-term memory (LSTM). However, neither PFNM nor FedMA can achieve ideal learning performance on non-IID data. 

\textbf{Discussion}. Since  model complexity can adapt to data, BNP methods can train local models more flexibly than parametric methods \cite{orbanz2010bayesian}. However, because the complexity of BNP models grows with an increase in data, BNP for FL raises the computational power of clients with a large amount of data.

\subsection{Server-side BFL}

Various server-side BFL methods implement the global aggregation or decomposition of all updated local models for clients in each communication round. We introduce Bayesian methods BME and BPD. Further, BFL works in other settings, such as continual learning using BCL.

\subsubsection{Bayesian Model Ensemble for FL Aggregation}

BL models approximate posterior distributions over model parameters using stochastic (e.g., MCMC) or deterministic (e.g., VI) \cite{kingma2013auto,wu2023evae} methods. For stochastic methods, the posterior distribution can be approximated by random sampling \cite{pml1Book}. Hence, each sample can be viewed as a base learner for the Bayesian model ensemble (BME).

In BME-based FL, FedBE \cite{chen2021fedbe} first utilizes the parameters of clients to construct a global posterior distribution (Gaussian or Dirichlet) in the aggregation phase of each communication round. MCMC then samples from this global posterior distribution to obtain the ensemble model, which is used on unlabeled data to obtain a pseudo-labeled dataset. Finally, a stochastic weighted averaging algorithm distills the global model on this pseudo-labeled data. Similarly, FedPPD introduces three aggregation algorithms with one serving  ensemble similar to FedBE. However, in the local learning phase of each communication round, FedBE still uses the point estimation method to obtain the local model parameters of clients, while FedPPD utilizes a BNN.

\textbf{Discussion}. BME helps to implement Bayesian inference on the server using the information of all clients more effectively to prevent model performance degradation. This works especially for local models on  non-IID data. Compared with other approximation methods, the sampling method is simpler and more accurate. However,  obtaining an accurate ensemble model requires as many samples as possible, increasing the computational power of devices. Also, we cannot obtain a specific global parameter directly by BME. Other methods for distilling an appropriate global parameter are required for client learning in the next communication round.

\subsubsection{Bayesian Posterior Decomposition for FL Decomposition}

For some ML tasks, it is essential yet challenging to decompose a model into a combination of sub-models that can be handled more easily than the original model \cite{abrial2009event}. This requires model decomposition. For Bayesian learning, Bayesian posterior decomposition (BPD) serves this purpose by decomposing the posterior of the global model to a combination of local models.

For BPD-based FL, by imposing strong constraints on uniform prior and local data independence, FedPA \cite{al-shedivat2021federated} decomposes global posterior distribution into the product of local posterior distributions of clients in each communication round. A concrete expression of parameters of the global posterior distribution can be obtained by federated least squares. Since the direct calculation of parameters from the global posterior will incur high computational and communication costs, calculating the global posterior is converted into an optimization problem and solved by sampling. Since the independent approximations of local posteriors cannot guarantee an accurate global posterior approximation, FedEP \cite{guo2023federated} extends FedPA to approximate a global model using an expectation propagation method. QLSD \cite{vono2022qlsd} uses the same method as in  FedPA to decompose the global distribution. QLSD mainly performs the update of  clients per the quantised Langevin stochastic dynamics and sends the resulting compressed gradient to the server to control communication consumption in FL. Differing from these models, FOLA also uses BPD to decompose the global posterior approximation into the product of local posterior distributions with weighted terms, which does not require any strong constraints. VIRTUAL \cite{corinzia2019variational} also uses BPD for aggregation on the server. To avoid catastrophic forgetting, its global posterior distribution is decomposed into the product of the global posterior distribution of the previous communication round and local posterior distributions (ratios of clients) of the current communication. 

\textbf{Discussion}. Most FL methods use naive parameter averaging FedAVG \cite{mcmahan2017communication} for model aggregation, which causes performance degradation when local data involves statistical heterogeneity. However, the stability of the BPD-based FL models on heterogeneous data can be significantly improved by global posterior model decomposition. More importantly, BPD also enables better interpretability of FL models. However, BPD introduces  challenges to FL. First, BPD may require other algorithms to assist model learning, which may incur a large computational overhead or even be intractable. Second, some decomposition methods may involve strong constraints, which are usually not feasible when solving practical problems.

\subsubsection{Bayesian Continual Learning for Continual FL}

Continual learning updates models over a variety of datasets, which may change over time in a sequential manner \cite{nguyen2018variational}. Bayesian learning can facilitate the property of continual learning and use the model posterior as the prior of the next task to achieve Bayesian continual learning (BCL).

For BCL-based FL, FOLA \cite{liu2021bayesian} uses the product of local posteriors to obtain a global posterior rather than a simple mixture of local posteriors for clients in each communication round for more robust learning on non-IID data. The resulting global posterior will then be sent to all clients at the next communication round as a prior. Similar to FOLA, in each communication round of pFedBayes, the prior distribution of a local model is replaced by the global distribution of the previous communication round to improve the learning performance and interpretability. Nevertheless, there are two main differences between FOLA and pFedBayes. First, the global distribution of pFedBayes is merely obtained by means of an incremental local model averaging, while the global distribution of FOLA is obtained by the product of local posteriors. Second, the local model for each client in pFedBayes is a BNN, while FOLA is a classic neural network.

\textbf{Discussion}. By applying BCL to FL, we can leverage the information aggregated from previous communication rounds. Such an online learning approach often results in better learning performance. However, a complex prior distribution may bring substantial computational overhead to modeling and result in limited performance improvement. It is thus challenging to arrive at a trade-off between using BCL and assuming a suitable prior distribution for FL.

\subsection{BFL from the Federated Perspective}

BFL research can also be categorized per other FL aspects as discussed in Sec. \ref{subsec:fl}. Accordingly, we review specific FL tasks using heterogeneous, personalized, hierarchical, dynamic and hybrid BFL methods. Further, federated Bayesian networks implement BL using FL or in a federated structure.

\textbf{Heterogeneous BFL} 
Heterogeneous BFL handles heterogeneous clients with different prior distributions, parameter distributions, or posterior distributions for individual clients. For example, \cite{kotelevskii2022fedpop} addresses heterogeneous clients using a mixed-effect model for each client, which includes fixed effect (common representation) and random effect (personalized representation). FedBE alleviates the heterogeneity between clients by fitting a posterior distribution for all possible global models.
However, existing research generally assumes all local models share the same global model architecture with minor adjustments on priors or parameter distributions. These cannot handle complex heterogeneities, such as heterogeneous structures and relations of clients \cite{li2020federated,Cao22d}.

\textbf{Personalized BFL}
Personalized BFL defines a personalized model for each client to deal with its distinct data distributions. For example, pFedBayes uses a personalized model for each client by minimizing the KL divergence between the global model and each updated local model. Instead, pFedGP constructs a personalized GP classifier for each client on its local data.
However, existing methods may focus on client heterogeneity but not personalize model parameters, learning tasks or objectives for each client \cite{tan2022towards}. 

\textbf{Hierarchical BFL}
Hierarchical BFL addresses scenarios where client features or edge nodes may be grouped into another layer, involving hierarchical FL structures. To the best of our knowledge, no work extends BL to hierarchical FL. In \cite{chen2022selfaware}, a hierarchical Bayesian model captures intra- and inter-client uncertainties to optimize hyperparameters: initial value, learning rate, and a number of (early-stop) steps. Nonetheless, existing research ignores complex interactions and couplings between client features or edge nodes, leading to poor performance \cite{briggs2020federated}. 

\textbf{Dynamic BFL}
Dynamic BFL handles evolving, temporal, or random characteristics, settings or tasks, which may also involves drifts (shifts, changes, variations) of client features, node features, or client-server interactions over time or other dimensions. These require dynamic BL mechanisms, such as updating priors, the number of clients, feature dynamics, and posterior contributions. To address the change in client contributions, the FBO for TFP uses BO to dynamically adjust the weights of clients in each communication round. FOLA uses BCL to dynamically adjust the prior distributions of clients to accelerate model convergence.
However, these methods generally overlook the dynamic components of FL, resulting in slow model convergence and a local optimal solution \cite{chen2021dynamic} and other gaps.

\textbf{Hybrid BFL}
Hybrid BFL is useful to address FL with mixed client structures, features, tasks, priors, posteriors, etc. In FedPPD, a BNN trains local models for FL. FedBE utilizes  a model ensemble to improve the robustness of FL.
While hybrid BFL models may be able to handle diverse problems, limited existing work is available \cite{huang2021starfl}. 

\textbf{Federated Bayesian networks}
In this review, we exclude research on federated Bayesian networks which use FL for BL. Examples are \cite{Ng022} which collectively learns a Bayesian network from partitioned data. Note that one may interchangeably use FBN for BFL, as in \cite{corinzia2019variational,KassabS22}. 

\textbf{Discussion}. The related research on these BFL topics is very limited, immature, incomplete and imbalanced. As discussed above, some of the settings, requirements or tasks have not been explored yet. In the following section, we further discuss these and the future directions of BFL research.

\section{Gaps and Directions}

BFL has demonstrated significant progress and potential for improving FL on limited, dynamic and uncertain data. However, the existing BFL research also shows significant limitations and gaps in addressing theoretical and practical FL requirements, problems, and challenges. 

\subsection{Gap Analyses}

Table \ref{Tab:BFL-comp} summarizes and compares both client- and server-based BFL methods discussed in Section \ref{sec:BFLtax-anal}. Their main limitations and gaps can be summarized as follows.

First, existing BFL methods have various limitations. Typical issues include high computational costs and low communication efficiency. Most BFL models involve strong constraints on (1) FL settings such as client independence and heterogeneity, privacy, resources, and communication costs, and (2) BL settings such as prior distributions and uncertainty of clients. These limit  BFL performance and applications. In addition, high computation and communication overheads limit the applications of BFL, particularly for decentralized, cross-device and personalized FL tasks. Their oversimplified settings and approaches thus lead to limited capacity and performance. 

Second, existing BFL exhibits weak-to-no capabilities in handling complex interactions and heterogeneities in FL applications with non-IID data. Real-life FL systems are non-IID, involving comprehensive non-IIDnesses \cite{Cao22d,cao2022beyond,Cao-rs-eng}. Examples include heterogeneous, interactive, coupled and hierarchical entities, features, relations, and structures within and between clients and communities and between clients and servers \cite{Cao22d}. Existing personalized FL and BFL only weaken these non-IIDnesses through strategies such as neutralizing heterogeneities across clients, e.g., by unified global optimal parameters or simplified Gaussian processes. The statistical heterogeneity between clients requires each client to learn its personalized optimal parameters. In addition, some devices and nodes may interact and couple with each other, while existing FL and BFL overlook their client couplings \cite{Cao22d,cao2022beyond,Cao-rs-eng,ijcai_PangCCL17}. In fact, most of existing references on non-IID FL do not address the above non-IIDnesses or significantly simplify these challenges \cite{Cao22d}.

In addition, BFL models still suffer from weak accuracy, robustness and learning performance. These may be attributed to poor-to-weak priors, heterogeneous, dynamic and hybrid clients or tasks, weak optimization, or inappropriate updating and aggregation mechanisms. There are also gaps in implementing local and global optimization in various settings, such as heterogeneous, hybrid, dynamic, and decentralized FL and for communication, updating, and aggregation. Regarding dynamic BFL, gaps exist in modeling evolving, drifting, nonstationary, or even unlimited scenarios over time or other dimensions (such as value domain, or state space). In summary, the limited research on BFL is at its early stage and focuses on simple applications of classic and main BL settings and methods in simplified FL scenarios, tasks or applications.

\begin{table*}
    \centering
        \caption{Categorization and comparison of various methods for Bayesian federated learning (BFL) from a Bayesian perspective.}
        \resizebox{.99\textwidth}{!}{
    \begin{tabular}{lrrrrr}
        \toprule
        Categories & BL & BFL models & Application & Advantages & Disadvantages \\
        \midrule
            & FBP & BDP \cite{TriastcynF19} & Medical images & Communication efficiency, high accuracy & Strong constraints \\
            & FBP & FDL-PP \cite{GuBP19} & Wireless communication & Low complexity, high accuracy & High computational cost, IID \\
            &  BNN & pFedBayes \cite{zhang2022personalized} & Finance, medicine & Limited data, non-IID & Strong constraints \\
            &  BNN & FedPPD \cite{bhatt2022bayesian} & Face perception, medical test & Weak constraints, non-IID & High computational cost\\
            & BO & FTS \cite{dai2020federated} & Human activity recognition & Communication efficiency, non-IID & Low convergence rate\\
            & BO & FBO for TFP \cite{zang2022traffic} & Traffic flow prediction & Communication efficiency, non-IID & Poor scalability\\
        Client-side   & BNP & pFedGP \cite{achituve2021personalized} & Health care, legal & Computational efficiency, non-IID & High computational cost\\
            & BNP & FedLoc \cite{yin2020fedloc} & Outdoor vehicle navigation & Data privacy, high accuracy & High computational cost, IID\\
            & BNP & FedCor \cite{tang2022fedcor} & Precision medicine & Communication efficiency, fast convergence &  Limited scenarios, IID\\
            & BNP & PFNM \cite{yurochkin2019bayesian} & Health care, finance & Communication efficiency & Limited scenarios, IID\\
            & BNP & FedMA \cite{wang2020federated} & Fingerprinting & Communication efficiency & High computational cost, IID\\
        \midrule
            & BME & FedBE \cite{chen2021fedbe} & Target localization & Deeper neural networks, non-IID & Strong constraints \\
            & BME & FedPPD \cite{bhatt2022bayesian} & Face perception, medical test & Model uncertainty, weak constraints, non-IID & High computational cost\\
            & BCL & FOLA \cite{liu2021bayesian} & Medical AI & Aggregation error, local forgetting, non-IID & High computational cost\\
        Server-side & BCL & pFedBayes \cite{zhang2022personalized} & Finance, medicine & Limited data, non-IID & Strong constraints\\
            & BPD & FedPA \cite{al-shedivat2021federated} & Pedestrian tracking & Computational efficiency, non-IID & Strong constraints\\
            & BPD & FedEP \cite{guo2023federated} & Disease detection & Communication efficiency, non-IID & Strong constraints\\
            & BPD & QLSD \cite{vono2022qlsd} & Autonomous driving & Communication efficiency, non-IID & Strong constraints\\
            & BPD & FOLA \cite{liu2021bayesian} & Medical AI & Aggregation error, local forgetting, non-IID & High computational cost\\
            & BPD & VIRTUAL \cite{corinzia2019variational} & Smart keyboards & Communication efficiency, non-IID & High computational cost\\
        \bottomrule
    \end{tabular}}
    \label{Tab:BFL-comp}
    \vspace{-10pt}
\end{table*}

\subsection{Research Directions}

On one hand, intricate real-world FL scenarios, requirement and applications challenge existing FL theories and systems, which could inspire promising BFL research issues and opportunities. On the other hand, BFL could transform FL and BL research and systems to a new generation. The new-generation BFL could include but may not be limited to: variational, non-IID, hierarchical, weakly-constrained, computation- and communication-efficient, hybrid, and actionable BFL theories and applications, and BFL under complex task, network and data conditions.  

\textbf{Variational FL}. Variational deep learning has made significant progress by integrating VI with deep neural learning \cite{wu2023evae}. 
Variational FL may expand the existing BFL research for (1) complex stochastic client/server conditions or settings; (2) diverse and efficient VI mechanisms in heterogeneous, hierarchical and hybrid FL settings; (3) large-scale and dynamic variational FL; and (4) variational FL under non-IID settings as discussed below. 

\textbf{Non-IID BFL}. Original BFL mechanisms, architectures and models are required to address non-IIDnesses \cite{cao2022beyond} in FL systems. Examples include (1) heterogeneous data structures, distributions, priors and parameterizations across clients, devices and nodes; (2) personalized client requirements, tasks and objectives, and sample/client importance difference; (3) interactive and coupled clients, communities and edge nodes, message sharing between clients; (4) interactions and couplings between clients and server; and (5) nonstationary, evolving, adaptive, and drifting client/server conditions. This goes beyond existing heterogeneous and personalized FL focusing on limited and neutralized heterogeneities of clients without client couplings.

\textbf{BFL on complex network conditions}. Real-life federated networks may involve complex settings or conditions. For example, new BFL theories may need to handle (1) disconnected nodes, unstably connected clients, siloed clients, and hybrid connections with online and offline devices; (2) unavailable, unstable or forgettable clients or nodes, or unseen domains; (3) coupled and interactive networks such as with cross-domain, cross-silo, cross-client or cross-node (client node or edge network) conditions; (4) unaligned or conflicting network conditions, such as with unaligned clients (where clients share inconsistencies such as on features, distributions, or communication), client conflict, domain conflict, or objective conflict (conflicts may exist in various real-life scenarios, e.g., contrary or inconsistent circumstances, distinct objectives, or different evaluation measures); (5) insecure networks such as with data or model poisoning attacks; (6) hierarchical networks, such as client clusters, multiple client communities, hierarchical centralized servers, and mixing decentralized and centralized client/server structures; and (7) mixing asynchronous and synchronous communications. 

\textbf{BFL on complex data characteristics}. FL applications often involve complex data, challenging the existing BFL and FL capacity. They include (1) low-quality data such as noisy, irregular and unaligned client data; (2) weak data such as with small, sparse and insufficient evidence; (3) changing data such as with covariate features, feature shift, prior shift, or concept drift; (4) mixed data such as with multiple domains, communities, modalities, structures and distributions, and multi-granular data such as with client and structure granularity; and (5) mixed labelled data with unlabelled to partially labelled clients. These require new BFL theories, e.g., for data-augmented, evolving BNP networks, and mixed-supervision BL networks and optimization for FL.


\textbf{Weakly-constrained BFL}. Existing FL and BFL involve various and often strong constraints, such as on client conditions, privacy, security, resources, client-server interactions, communication, prior and posterior distribution, parameter and message sharing, and optimization. These conflict with the diversified FL reality, where there may be siloed and coupled clients, some client features may be unbalanced and sparse, and data may be mixed, evolving and hybridized. These require substantially new and more flexible BFL theories and models catering for specific requirements. 

\textbf{Computation- and communication-efficient BFL}. Both existing FL and BFL models face the significant challenges of high computation and communication costs and low efficiency, although intensive efforts have been made to address these. Computation- and communication-efficient BFL requires more efficient, energy-aware and scalable learning theories, sampling, optimization methods, client updating, server aggregation, parameter and message sharing, and back-propagation mechanisms.

\textbf{Hybrid BFL}. Real-life FL applications may involve (1) mixed features, modalities, and data sources; (2) hybrid priors and distributions of client data; (3) multiple to hybrid FL tasks; (4) mixing centralized and decentralized clients/server; and (5) mixing BFL with learning paradigms, etc. These require new BFL theories, architectures, and mechanisms for hybrid Bayesian learning, multi-task BFL, multi-source BFL, multimodal BFL, hybrid FL client updating, server pooling and aggregation, client-server interaction, and optimization methods. Other hybrid BFL areas include hybridizing BFL (1) with other learning systems, such as for Bayesian federated transfer learning, reinforced BFL, ensemble BFL, and BFL for anomaly detection; and (2) with various communication and computing settings, such as compressed BFL, encrypted BFL, asynchronous BFL, blockchained BFL, decentralized BFL, Bayesian federated edge learning, over-the-air BFL, sparse BFL, adaptive BFL, BFL with message sharing across devices or communities, and cross-silo BFL. 

\textbf{Actionable BFL}. Actionable BFL requires extra functional and nonfunctional requirements, settings or performance to ensure the actionability \cite{pakdd_CaoZ06,widm_Cao13,Cao22j} of FL systems in the real world. This requires BFL theories and systems to support functions or capabilities such as fairness, unbiasedness, robustness, resilience, security, safety, responsibility, verifiability, explainability, and ethics. The evaluation of BFL algorithms and systems would have to consider both technical (such as statistical significance) and domain-driven business-oriented (such as impact on business) aspects and measures, and possibly both objective and subjective evaluation measures driven by the domain knowledge and factors, etc. \cite{dddm_Cao10}.  

Here, we only address a few opportunities that could directly or naturally benefit from addressing intrinsic FL issues, utilizing stronger BL mechanisms, or aiming for better capacity and performance. In fact, more exciting opportunities exist in exploring intricate FL demand and challenges in the real world, and focusing on seamlessly integrating BL theories into FL settings. 

\section{Conclusion}

By applying Bayesian learning  to the federated learning framework, Bayesian federated learning has become an important learning paradigm to handle various FL challenges and requirements for more robust uncertainty learning. While existing BFL methods exhibit significant progress and potential in learning with limited data and uncertainties, various technical gaps and hence opportunities remain. Fundamental BFL research is required to handle stochastic, heterogeneous, nonstationary, interactive, hierarchical, imbalanced, unlabeled, personalized, and hybrid challenges and requirements for robust BFL theories and actionable FL applications. 


\bibliographystyle{named}
\bibliography{ijcai23}

\end{document}